\documentclass[letterpaper]{article} 
\usepackage{aaai18}  
\pdfoutput=1

\usepackage{times}  
\usepackage{helvet}  
\usepackage{courier}  
\usepackage{url}  
\usepackage{graphicx}  
\usepackage{amsmath}
\usepackage{multirow}

\usepackage{smartdiagram}
\usesmartdiagramlibrary{additions}
\smartdiagramset{back arrow disabled=true,
  uniform color list=white for 6 items,
  uniform arrow color=true,
  arrow color=black,
  arrow tip=latex,
  arrow line width=2pt,
  module minimum width=2.75cm,
  text width=2.5cm,
  module x sep=3.65cm,
  additions={
    additional item offset=0.75cm,
    additional item border color=gray,
    additional connections disabled=false,
    additional arrow color=black,
    additional arrow tip=stealth,
    additional arrow line width=2pt,
    additional arrow style=-latex,
    additional item width=2.75cm,
    additional item text width=2.65cm,
    additional item shadow=drop shadow,
    additional item font=\small
}}

\usepackage{tikz}
\usetikzlibrary{shapes,arrows}
\usetikzlibrary{shadows.blur}

\nocopyright 

\begin{document}

\title{Word Vector Enrichment of Low Frequency Words in the Bag-of-Words Model for Short Text Multi-class Classification Problems}
\author{Bradford Heap\textsuperscript{1},
Michael Bain\textsuperscript{1},
Wayne Wobcke\textsuperscript{1},
Alfred Krzywicki\textsuperscript{1} \and
Susanne Schmeidl\textsuperscript{2} \\
\textsuperscript{1}School of Computer Science and Engineering\\
\textsuperscript{2}School of Social Sciences \\
The University of New South Wales\\
Sydney NSW 2052, Australia \\
\texttt{\{b.heap,m.bain,w.wobcke,alfredk,s.schmeidl\}@unsw.edu.au}
}
\maketitle

\setcounter{footnote}{0}

\begin{abstract}
The bag-of-words model is a standard representation of text for many
linear classifier learners.
In many problem domains, linear classifiers are preferred over more
complex models due to their efficiency, robustness and interpretability,
and the bag-of-words text representation can capture sufficient information
for linear classifiers to make highly accurate predictions.
However in settings where there is a large vocabulary, large variance in
the frequency of terms in the training corpus, many classes and very short
text (e.g., single sentences or document titles) the bag-of-words
representation becomes extremely sparse, and this can reduce the accuracy of
classifiers.
A particular issue in such settings is that short texts tend to contain
infrequently occurring or rare terms which lack class-conditional evidence.
In this work we introduce a method for enriching the bag-of-words model by
complementing such rare term information with related terms from both
general and domain-specific Word Vector models.
By reducing sparseness in the bag-of-words models, our enrichment approach
achieves improved classification over several baseline
classifiers in
a variety of text classification problems. 
Our approach is also efficient because it requires no change to the linear
classifier before or during training, since bag-of-words enrichment applies
only to text being classified.
\end{abstract}

\section{Introduction}
The bag-of-words model is used as the standard representation of text input for many linear classification models (such as Multinomial Naive Bayes \cite{mccallum1998comparison} and Support Vector Machines \cite{Joachims98}).
Linear classifiers have been widely studied for many text classification
problems~\cite{Seba:j:2002}, such as document classification or sentiment
analysis, due to their efficiency, robustness and
interpretability, and the bag-of-words text
representation can capture sufficient information for linear classifiers to
make highly accurate predictions~\cite{DumaisPHS98}.
However, in settings where there is a large vocabulary, a high number of classes (e.g., complex ontologies) 
and short text (e.g., fragments of text, single sentences or document titles) the bag-of-words representation contains extremely sparse data
which reduces the accuracy of the linear classification models \cite{Wang:2012}.

A particular problem with classification of short texts 
is low frequency (rare) words, or as an extreme, words that do not occur at all in the text
used to train models, but do occur in test data. We have found that, despite the large volumes of text available on the Internet, low frequency words
present a problem for classification in three different domains that are investigated in this paper: (i) Reuters news article classification,
(ii) classification of journal article titles into medical ontologies, and (iii) classification of text snippets into ``conflict drivers'' for use in social science
conflict analysis. These classification problems are also characterized by having a large number of categories (typically around 60 classes at a minimum),
and a highly imbalanced distribution of instances over the classes. As there is limited data in the smaller classes, many supervised text classification
methods are less accurate for these classes \cite{joachims1997probabilistic,RennieSTK03}.
In addition, due to the specialized technical vocabulary used in the
medical and social science fields, such classification problems are
influenced by expert background knowledge not directly expressed in the
text itself~\cite{GabrilovichM06}.
Moreover, in conflict analysis, the classification of events
(to identify their significance in relation to a conflict) from official
reports in which details are aggregated over time and location means that
superficially similar texts may have very different
classifications~\cite{Donn:etal:c:2014}.

In this work, we present an approach for increasing the predictive power of low frequency and out of vocabulary words by enriching the bag-of-words model with related terms found in a Word Vector model, trained on either a general or domain-specific corpus. 
Word Vector models build on the ideas of capturing word co-occurrence information 
and assume that contextually related words will often occur with a similar set of surrounding words \cite{BengioDVJ03}. 
These models are constructed using neural networks and project words into a dense vector space. 
Two words that are close to each other in this space, either in distance or cosine angle, often have semantic or syntatic similarity \cite{CollobertW08,abs-1301-3781}. 

Our method of bag-of-words enrichment assumes that words that occur rarely in the training data may have neighbours in the Word Vector space which are able to aid in the classification of the short text. 
Using the Word Vector model, we enrich the bag-of-words representation by including neighbouring words found in the Word Vector model. 
Capturing related terms allows our approach to increase the number of features which a linear model uses to make a classification prediction.
Importantly, this approach can handle unseen words efficiently because it does not require retraining of the classification models during testing,
as enrichment applies only to the text to be classified. 

Our empirical evaluation shows that this method produces increased classification accuracy in three domains.  
The first dataset used for evaluation is a subset of the benchmark Reuters-21578 dataset which considers the classification of short financial news articles. 
The second evaluation is performed on a subset of the benchmark OHSUMED dataset \cite{HershBLH94} in which we consider the classification of medical articles which contain a title but no abstract. 
The third evaluation is performence on a dataset which classifies conflict drivers extracted from NGO analyst reports on the
Democratic Republic of the Congo (DRC).
In the light of the results, we explore the tradeoffs between the use of general and domain-specific corpora for training: general corpora with the advantage
of a large amount of training data, but lacking domain-specific technical
terms, and domain-specific corpora lacking quantity of data.

The paper is organized as follows. Starting with a brief description of the bag-of-words model and its limitations, and presenting various Word Vector models,
the main part of the paper is a description of our approach to Word Vector enrichment of the bag-of words model, and the empirical evaluation. We conclude
with a discussion of related work.


\section{Limitations of the Bag-of-Words Model}
We define a text \(T\) to be a sequence of individual word tokens \(T = [t_1, t_2, \ldots]\).
The bag-of-words model maps each text \(T\) to a large vector \(M_T\) of dimension \(|V|\), where
the vocabulary \(V\) is the set of all tokens in a corpus of training data. 
In our setting, this labelled training data is an assignment of a single class label to each instance of a training set. 

The bag-of-words vector \(M_T\) contains the term frequency of each token from \(V\) that occurs in \(T\), where the value of the vector is 0
for every token in \(V\) that does not occur in \(T\) \cite{Scott98textclassification,WangD08}. Of course, if \(T\) contains a token that does not
occur in the vocabulary of the training corpus, there is no element of the vector corresponding to that token.
Thus the bag-of-words model assumes that the presence and frequencies 
of tokens are important to the classification of text and the ordering of the words is not.

Although the bag-of-words model is widely used and performs exceptionally well in many problem domains, it contains a number of limitations which are particularly problematic in short text classification problems. 
In our problem domain there are four key issues: 

\subsubsection{Vocabulary Size.}
The dimensionality of the bag-of-words vector is the number of distinct words in the vocabulary. 
For Support Vector Machines (SVM) this can result in a large training time due to the interaction between training instances and the dimensionality of the bag-of-words vector.
In particular, as the number of training instances increases the number of unique words in the vocabulary is likely to increase.
This in turn, then increases the dimensionality of the bag-of-words vector and, as the optimization to find the SVM hyperplane needs to be calculated across every dimension, the training time increases. 
In the Multinomial Naive Bayes (MNB) classifier, the size of the vocabulary is used to calculate each term's conditional probability.  
This produces very small conditional probabilities and can be particularly problematic in classes with a small class vocabulary or a small number of training examples.

\subsubsection{Imbalanced Class Distribution.}
For the problems of short text classification that we have examined, the class distribution is highly uneven. This in turn biases many classifiers to prefer
large classes, performing poorly on (the larger number) of smaller classes. 


\subsubsection{Rare and Out of Vocabulary Words}
The frequency distribution of word usage in typical text is extremely skewed (following Zipf's law). 
Words that occur only very infrequently do not give a learning algorithm sufficient information to determine their correct influence on classification.
Words that do not occur in training data are discarded by the bag-of-words model as there is no knowledge of their classification.


\subsubsection{Term Overlap Between Classes.}
In problems such as sentiment analysis or topic classification where there is a clear distinction between terms pertaining to certain classes,
linear models using the bag-of-words representation work well. 
However, in problems with a large overlap between the terms used to describe two or more classes, accuracy can substantially decrease \cite{Scott98textclassification}. 
For example, in social science conflict analysis, the words used in news reports classified as ``condemning an attack'' have much common
with those labelled ``military attack''. 

\begin{figure*}
\centering
\tikzstyle{block} = [rectangle, draw, fill=white,  
    text width=7.5em, text centered, rounded corners, minimum height=4em, blur shadow={shadow blur steps=10}, node distance=3.5cm]
\tikzstyle{line} = [draw, -latex, line width=2pt]

\small
\scalebox{0.84}{    
\begin{tikzpicture}[node distance = 2cm, auto]
    \node [block] (init) {Input: \\Instance Bag-of-words Vector};
    \node [block, right of=init, yshift=25pt] (identify) {2) Create Neighbouring Bag-of-words Vectors for each Rare Word};
    \node [block, right of=identify] (evaluate) {3) Aggregate into an Enriched Bag-of-words Vector};
    \node [block, above of=init, node distance=1.5cm] (update) {1) Input: \\Dense Word Vector Model};
    \node [block, right of=evaluate] (decide) {4) Classification Model uses Enriched Bag-of-words Vector};
    \node [block, right of=decide] (stop) {Output: Classification Prediction for Instance};
    \path [line] (init) -- (identify);
    \path [line] (update) -- (identify);
    \path [line] (identify) -- (evaluate);
    \draw [-latex, line width=2pt] ([yshift=-7pt] init.east) -- ([yshift=-5pt] identify.south) -| (evaluate.south); 
    \path [line] (evaluate) -- (decide);
    \path [line] (decide) -- (stop);
\end{tikzpicture}
}
\caption{Bag-of-Words enrichment with Word Vectors Process}
\label{fig:process}
\end{figure*}
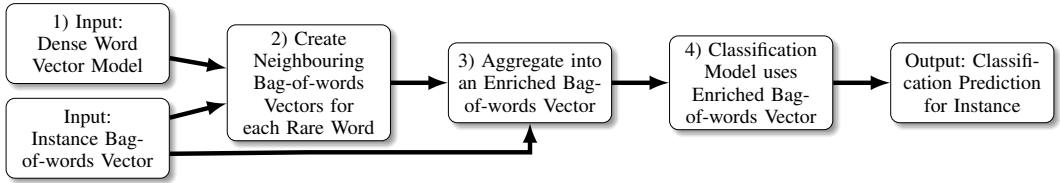

\section{Word Vector Models}
Word co-occurrence information can be used to build models that capture the relations between words generally not captured in dictionary-based enrichment systems \cite{mandala1998use}. 
For example the proper noun ``May'' and the title ``Prime Minister'' co-occur with high frequency in text related to the United Kingdom, but outside of this domain these two terms may not be as strongly related.
Although co-occurrence models are able to capture relations in the same sentence they fail to capture synonymous terms, for example, the words ``tumor'' and ``tumour'' \cite{mandala1998use}. 

Word Vector models are an extension of the ideas of word co-occurrences 
and build on the idea that contextually related words often occur with a similar set of surrounding words \cite{BengioDVJ03}. 
Word Vector models are constructed by projecting all the words from a corpus into a dense vector space,
for example, to reduce a bag-of-words vector with a dimension of 100,000 to a 100-dimensional ``distributed feature vector''. 
Two words that are close to each other in this space, either in distance or cosine angle, often share semantic or syntactic similarity \cite{CollobertW08,abs-1301-3781}. 

Word Vector models are constructed using neural networks  \cite{BengioDVJ03,CollobertW08}. 
Construction of the models is unsupervised and generally involves huge corpora of documents (e.g., Wikipedia \cite{CollobertW08} or Google News \cite{abs-1301-3781}). 
In the training phase, the problem is typically to predict the next word in the text using the input of the previous \(n\) words in a sentence or document \cite{BengioDVJ03,LeM14}. 
Word similarity is achieved in the intermediate layers of the neural network as a by-product.
Combining Word Vector models with other machine learning approaches has shown promising results.  
In particular, Maas et al. \shortcite{MaasDPHNP11} used Word Vector
models to discover additional features for sentiment analysis using SVM.

\section{Bag-of-Words Enrichment with Word Vectors}

In this section we describe our approach for enriching the bag-of-words vector by leveraging information from contextually related words found in an unsupervised Word Vector model. 
This allows the bag-of-words model to contain more non-zero elements which provides more information to the linear classifiers for making predictions. 

This process of enriching the bag-of-words model involves a number of steps which are shown in Fig. \ref{fig:process} and described below:
\begin{enumerate}
 \item Train or obtain an unsupervised Word Vector model \(W\) with dimension \(d\) that models the vocabulary and language of the sentences in the labelled or
 unlabelled data and/or any other domain related text.
 The Word Vector model must, for a given token \(t\) in the vocabulary, output a list of tokens related to \(t\), ranked in order of similarity. 
 \item For an unlabelled short text \(T\) and each token \(t_i \in T\) with a frequency in the labelled training data less than \(n\) (i.e., \(t_i\) is a rare word), use the Word Vector model to find the \(k\) nearest neighbours of \(t_i\) which occur with any frequency greater than zero
 in the labelled training data and
 construct a new bag-of-words vector \(M_{t_i}\) with these up to \(k\) elements assigned a term frequency of \(1\) and all other elements in the vector assigned the value \(0\).
  \item From the original bag-of-words vector \(M_T\), create
  an enriched bag-of-words vector \(M_{T^+}\) by adding 
  \(M_T\) and each individual rare token's nearest neighbour bag-of-words vector \(M_{t_i}\).
  \begin{equation}
  M_{T^+} = M_T + \sum M_{t_i}   
  \end{equation}
\end{enumerate}

As a result of this process, \(T\) is now classified
using the enriched bag-of-words vector \(M_{T^+}\) in lieu of the original bag-of-words vector \(M_T\). 

We now work through each of these steps showing how the Word Vector model can be trained to capture contextually similar words and how this is particularly beneficial for words with low frequency in the labelled training data. 
We are then able to show how enriched bag-of-words vector harnesses this additional information to address the previously outlined shortcomings of the bag-of-words model. 

\subsection{1. Word Vector Training Algorithms}

In this work, we train Word Vector models using Mikolov et al.'s \shortcite{abs-1301-3781} Continuous Skip-gram Model as it addresses our requirement that the Word Vector Model captures both syntactically related and common co-occurring terms. 
This training algorithm differs from previous word embedding models, 
instead of using the previous and next \(n\) words to predict a missing word, only one word is used as input and the task is to predict the words which occur in sentences around the given input word.
This approach captures the relationships between both co-occurring words and syntactically related words. 
For example the synonymous terms ``Daesh'' and ``ISIS'' will both produce related words of ``fighters'', ``terror'' and ``loyalists'' as in the skip-gram training model they will appear in a very similar area of the vector space. 
Furthermore, in a domain specific model the term ``Trump'' will produce the common co-occurring word ``President''. 
The technical details and empirical evaluations of this approach have been presented in previous works \cite{abs-1301-3781,MikolovSCCD13}.

\subsection{2. Constructing Nearest Neighbour Bag-of-Words Models using Word Vectors}
Once trained, the distance between individual tokens in a Word Vector model is agnostic to the distribution of word frequencies. 
This means that any two neighbouring terms in the Word Vector model share similarity with the words which they co-occurred with during the construction of the model independent of the number of training examples for each term. 
This can result in rare terms having high frequency terms as close neighbours. 
This close mapping between related terms allows us to find more frequently occurring terms for a given rare term. 
By capturing these more common related terms we can use their information in the model to aid in the classification of a short text. 

The construction of a nearest neighbour bag-of-word model requires two hyper-parameters:
the rare word frequency threshold \(n\) 
and the number of neighbouring words \(k\) to include from the Word Vector model. 
Suitable values for these parameters needs to be established through empirical evaluations (e.g., a grid search), however, both of these values depend on the labelled training data and not on the Word Vector model,
and this allows the Word Vector model to be trained in 
isolation to the supervised model. 
Furthermore, as the Word Vector model training algorithm has no dependencies between training instances, the Word Vector model can be updated as new test instances are supplied to the model. 
This allows the Word Vector model to capture the relationships between words which may be out of the vocabulary in the model and map them to other words for which there is training data. 
This capturing of out of vocabulary words addresses one of the major issues with the classical bag-of-words model. 

\subsection{3. Aggregation of the Nearest Neighbour Bag-of-Words Vector}
As a final step before classifying the short text, we combine the original bag-of-words vector with the bag-of-word vectors constructed for each of the rare words in the original short text. 
This aggregation of the bag-of-word vectors has three benefits:
1) reinforcement of the term frequencies of any word which occurs in the original bag-of-words and
then is a nearest neighbour of a rare word;
2) an increase in the number of non-zero elements in the enriched bag-of-words vector through the inclusion of neighbouring words; 
3) out of vocabulary words (not contained in the labelled text) are able to contribute to the classification through their nearest neighbours found in the Word Vector model (when the Word Vector is trained on a larger corpus).

This approach addresses the classical bag-of-words model problems of: 
a) rare and out vocabulary words, and 
b) term overlap between classes as the combination of multiple related words should have a much stronger affinity to a narrow set of classes. 

As an example, 
consider the classification of the following sentence ``Officials said that supporters of Mullah Haibatullah Akhundzada's faction and Mullah Rasoul's supporters clashed in Shindand district of the province''.
Without any other context it is hard to determine if this clash is between rival political parties or militant groups (the terms ``officials'', ``supporters'', and ``faction'' commonly describe both).
To illustrate how word vector enrichment can be used to classify this sentence,
suppose the name ``Haibatullah'' is a rare term (it only occurs three times in the training data).
Using a domain specific Word Vector model, the neighbouring words of ``Haibatullah'' are ``Mullah'' and ``Omar''.
The bag-of-words vector for the original sentence is then enriched with these terms.
As ``Mullah Omar'' is a commonly occurring name in the training data,\footnote{Haibatullah is the current leader of the Taliban (since May 2016) and Mullah Omar is a former leader.} 
the classifier has more information to 
determine the correct label of the sentence.

\section{Short Text Classification Models}

Linear classification models are able to use the bag-of-words model to make predictions about the classification of text. 
In this work we focus on two widely used classification models, Multinomial Naive Bayes (MNB) and Support Vector Machines (SVM). 
These two models have been widely established as baselines methods for text classification across many domains \cite{Wang:2012}. 

\subsubsection{Multinomial Naive Bayes.}
These classifiers apply the Bayes rule over a collection of labelled training examples to estimate the probability of a given text belonging to a particular class. 
Formally the classifier is defined as:
\begin{equation}
 C_{\textrm{NB}} = \underset{c\in C}{\textrm{argmax}}P(c) \underset{t_i \in T}{\mathbf{\Pi}} P(t_i|c)
\end{equation}
where \(P(c)\), the prior probability, is the probability of a given class out of the set of all classes in the training set, without any other knowledge about the contents of the text to be classified.
The conditional probability of a given token for a particular class \(P(t_i|c)\) is calculated using Laplace smoothing to avoid zero probabilities. 

\subsubsection{Support Vector Machines.}
These classification models are trained to find a hyperplane separating two classes in a high dimensional space.
The classification of an unlabelled text is then decided by which side of the hyperplane the features (tokens) of the text fall. 
%
In problem domains with many class labels, a single class classification requires the construction of SVM models for every pair of classes.
A text is then classified by every classification pair and the class which the text is classified into most often 
\cite{HastieT97}.

It is straightforward to see how both the MNB and SVM classifiers are to use this enriched bag-of-words model to make classification decisions which utilize more of the training information. 
In particular, the generative MNB model is now able to use more non-zero elements of the bag-of-words vector to calculate a classification probability. 
The discriminative SVM model is able to use the enriched bag-of-words vector to more confidently decide which side of the hyperplane the short text should be classified on. 

\subsection{Neural Network Classifiers}
We compare the linear classifiers with two recently developed neural network models for text classification which use data from a pretrained Word Vector
as input into a further neural network model.

\subsubsection{Paragraph Vectors.}
Paragraph Vectors (also called \emph{doc2vec}) are an extension of Mikolov et al.'s \shortcite{abs-1301-3781} \emph{word2vec} word embedding training algorithm.
The Paragraph Vectors model is designed to overcome the problems of the loss of word ordering information in bag-of-words models \cite{LeM14}. 
The primary change is during training class label features (``Paragraph IDs'') are inserted into the same dense vector space as the words.
After training, classification of text is performed by calculating the mean vector of the words in a text to be classified, and then
the text is labelled with the Paragraph ID feature nearest to the mean vector. 
Recent analysis has shown that replicating the results of the original Paragraph Vectors paper has been difficult due to the nature of tuning the model's hyperparameters \cite{LauB16}.

\subsubsection{Convolutional Neural Networks.}
Kim \shortcite{Kim14} uses pre-trained Word Vector models as input into a one layer Convolutional Neural Network (CNN). 
This approach learns sequences of words which is otherwise lost in using the standard bag-of-words model. 
This work showed that a CNN could be used to classify text in sentiment analysis and question classification benchmark problems. 
However, subsequent work has shown that tuning the large number of hyperparameters is difficult and often impractical due to the computational cost \cite{ZhangW15b}. 

For the neural network based Paragraph Vector model, including additional words during classification will change the mean word vector for a given input sentence. 
As this change only affects the text to be classified, it is possible that the nearest Paragraph ID to the mean may differ from that of the original text. 
Because the two dimensional structural patterns identified by the CNN are attuned to word adjacency in the input
inserting additional words into the middle of the text to be classified will 
have a detrimental effect on the recognition of such patterns.
Due to the behaviour of the training and classification algorithms of both neural network based classifiers our empirical evaluations show that our enrichment method does
not improve the classification accuracy for these neural network models. 

\section{Evaluation}
Our evaluation over the three datasets is setup as a 10x10-fold cross validation. 
For each dataset we train a domain specific Word Vector model and use it for enrichment in each evaluation fold.\footnote{We use DeepLearning4J's implementation of Mikolov's Continuous Skip-gram algorithm (https://deeplearning4j.org/).}
We set the hyper-parameters of each Word Vector model to a dimension \(d = 100\), a training window size of 10, a word frequency of at least two and train for 10 epochs. 
We train two linear model classifiers, the first is the standard MNB model and
the second is the SVM model which we train using Weka's\footnote{http://www.cs.waikato.ac.nz/ml/weka/} implementation of the SMO training algorithm (with hyper-parameter \(C=1\)). 
We use DeepLearning4J's implementation of Paragraph Vectors and Kim's CNN to train our neural network based classifiers.
All classification models are configured to always produce a single classification prediction for a given input.

Our evaluation setting is a single label, multi-class classification problem where every test instance is assigned one class label.
In this setting, recall is a suitable metric to measure the performance of our algorithms. Suppose there are \(m\) classes \(C_1, \cdots, C_m\),
and that \(C = C_1 \cup \cdots \cup C_m\) is the set of all instances.
The \emph{recall} of a single class \(C_i\) (class recall) is defined as the number of correct classifications from $C_i$ (true positives)
for this class \(tp_i\) divided by the number of instances in the class \(|C_i|\), assumed non-zero. Then \emph{micro recall} is defined
as the recall over the aggregated set of classes, or equivalently, as a weighted sum of the class recall measures, as follows:
\begin{equation}
 \mbox{micro recall} =  \frac{\sum^{m}_{i=1} tp_i}{|C|} =  \sum^{m}_{i=1} \frac{tp_i}{|C_i|} \frac{|C_i|}{|C|}
\end{equation}
Note that recall is equal to precision (and F1) in this setting because there is no need to distinguish the relevant and irrelevant instances
over the whole set; more precisely, every instance of $C_i$ that is not classified correctly 
(a false negative for $C_i$) is a misclassification
in another class (a false positive for that class). Similarly, micro recall is equal to accuracy.

The reason to focus on recall is that, as our approach is designed to enrich classes with few training examples, we measure this effect using macro recall,
defined as the mean of the individual class recall measures:
\begin{equation}
 \mbox{macro recall} =  \frac{1}{m} \sum^{m}_{i=1} \frac{tp_i}{|C_i|}
\end{equation}
In this metric, each class recall is weighted equally in the calculation; with micro recall, each class recall is weighted by the proportion of
instances in that class (as in the second formulation above), hence favours the larger classes.

For each dataset and classification model we tune the hyper-parameters for the label frequency \(n\) and the number of additional words from the 
Word Vector model \(k\) using a grid search over the first fold of the first evaluation set.  
Finally, 
we use the non-parametric Wilcoxon Signed-Ranked test for statistical significance to measure the benefit of our method in the linear classifiers. 

\begin{table*}[t] 
\centering \small
\caption{\small Reuters-21578 Subset (3,003 instances, 93 Classes): Domain Specific Word Vector.}
\label{tab:reutersrecall}
\begin{tabular}{|l|c|c||r|r|r||r|r|r|}
\hline
\multicolumn{3}{|l||}{} & \multicolumn{3}{c||}{\textbf{Micro Recall}} & \multicolumn{3}{c|}{\textbf{Macro Recall}} \\ \hline 
 & \(n\) & \(k\) & \multicolumn{1}{c|}{Baseline} & \multicolumn{1}{c|}{WV Enrichment} & \multicolumn{1}{c||}{Error Reduction} & \multicolumn{1}{c|}{Baseline} & \multicolumn{1}{c|}{WV Enrichment} & \multicolumn{1}{c|}{Error Reduction} \\ \hline  
MNB                                      & 3 & 3 & 0.765                      & 0.785                      & 8.75\%                       & 0.178               & 0.212         & 4.14\%  \\ \hline
SVM                                      & 3 & 3 & 0.842                      & 0.854                      & 7.66\%                       & 0.443               & 0.475         & 5.75\%  \\ \hline 
Paragraph Vectors                        & 3 & 3 & 0.795                      & 0.786                      & -4.71\%                      & 0.443               & 0.446         & 0.62\%  \\ \hline 
CNN                                      & 3 & 3 & 0.775                      & 0.759                      & -7.38\%                      & 0.324               & 0.311         & -1.86\% \\ \hline
\end{tabular}
%
\centering \small
\caption{\small Reuters-21578 Subset (3,003 instances, 93 Classes): Google News Word Vector.}
\label{tab:reutersrecall2}
\begin{tabular}{|l|c|c||r|r|r||r|r|r|}
\hline
\multicolumn{3}{|l||}{} & \multicolumn{3}{c||}{\textbf{Micro Recall}} & \multicolumn{3}{c|}{\textbf{Macro Recall}} \\ \hline 
 & \(n\) & \(k\) & \multicolumn{1}{c|}{Baseline} & \multicolumn{1}{c|}{WV Enrichment} & \multicolumn{1}{c||}{Error Reduction} & \multicolumn{1}{c|}{Baseline} & \multicolumn{1}{c|}{WV Enrichment} & \multicolumn{1}{c|}{Error Reduction} \\ \hline  
MNB                                      & 3 & 3 & 0.765                      & 0.773                      & 3.49\%                       & 0.178               & 0.186         & 0.97\%  \\ \hline
SVM                                      & 3 & 3 & 0.842                      & 0.843                      & 0.84\%                       & 0.443               & 0.449         & 1.08\%  \\ \hline 
\end{tabular}
\end{table*}

\subsubsection{Reuters-21578 Dataset}
This corpus contains news reports from 1987 and is a benchmark dataset widley used in many previous text categorisation evaluations. 
For our evaluation on short text 
we extract a subset of the reports where the length of the article body is 100 words or fewer (approximately two sentences). 
We exclude any article belonging to the `earn' class as these documents are share ticker information and do not form full sentences. 
This subset dataset contains 3,003 articles labelled by 93 categories.
The largest class `acq' contains 1,376 documents, followed by the `money-fx' class with 291 documents, 73 classes have fewer than 30 documents. 
This subset dataset contains 7,213 unique words and the mean length of each document is 70 words. 
We train the Word Vector model used for enrichment over the full Reuters-21578 corpus.

Table \ref{tab:reutersrecall} shows the results of our evaluation using the full document text without any pre-processing. 
These results show that at the baseline level the SVM classifier produces the highest recall.  
The MNB results are the lowest across all four classification models, this is likely a result of the skew in the data having a large influence on the 
prior probability component of the MNB model and therefore it is more likely to make classifications into the largest classes. 
These results show that using a Word Vector model to enrich the bag-of-words representation improves both micro and macro recall results of the MNB and SVM classifiers. 
For both of these linear classifiers the Wilcoxon Signed-Ranked test shows statistical significance of \(p < 0.000005\) in improving recall. 

In contrast the neural network classifiers have reduced micro recall and almost no change in macro recall with enrichment. 
As discussed earlier, the lack of improvement for these classification algorithms through the inclusion of additional words is likely related
to the structural representation of the text
as a single geometric mean of all the words used in the Paragraph Vector model 
and word ordering patterns in the CNN model.
These approaches assume the structure of the input data is very similar to their training data and their treatment of the sequence of words is very different to 
the linear classifiers which treat each word independently. 

Due to the complexity of the neural network models there is a large difference in the time required to train and evaluate each classification algorithm. 
The Word Vector model used for enrichment across all the models takes 10 minutes to train. 
The quickest classification model is the MNB model which performs the full 10x10-fold cross validation in under 10 minutes. 
The SVM model takes three hours to build a model and evaluate each fold. 
The neural network classifiers are much slower, Paragraph Vectors requires a day to train and evaluate and the CNN model takes three days. 
Furthermore, a large amount of time is required to tune the hyper-parameters of the Paragraph Vectors and CNN models which are both very sensitive to small changes.
Although some of this variance in performance is explained by the implementation of each model, 
due to the computational time and the lack of improvement using Paragraph Vector and CNN models we limit further evaluation to only the linear 
classifiers.

In the above evaluation we trained a domain specific Word Vector model using the full Reuters-21578 dataset.
The benefit of the domain specific Word Vector model is that it should be expected that words which have close domain specific similarities 
are modelled better than a general model trained over a much larger corpus of text. 
To evaluate this hypothesis we performed a second evaluation of the Reuters dataset which used Mikolov's Google News pre-trained Word Vector model for enrichment. 
This pre-trained model was trained over 100 billion English words containing three million unique words and phrases.
Table \ref{tab:reutersrecall2} presents the results of this evaluation for the linear classifiers. 
These results show that the large general Google News model can also be used with our algorithm to enrich Bag-of-words vectors and improve the recall of these classifiers.
Although these results are statistically significant \((p < 0.05)\), the error reduction is much smaller than the previous results, 
which is likely a result of the additional words 
having less relevance to the 
financial news domain. 

\begin{table*}[t] 
\centering \small
\caption{\small OHSUMED Subset (12,752 instances, 21 Classes): Domain Specific Word Vector.}
\label{tab:ohsumedfull}

\begin{tabular}{|l|c|c||r|r|r||r|r|r|}
\hline
\multicolumn{3}{|l||}{} & \multicolumn{3}{c||}{\textbf{Micro Recall}} & \multicolumn{3}{c|}{\textbf{Macro Recall}} \\ \hline 
 & \(n\) & \(k\) & \multicolumn{1}{c|}{Baseline} & \multicolumn{1}{c|}{WV Enrichment} & \multicolumn{1}{c||}{Error Reduction} & \multicolumn{1}{c|}{Baseline} & \multicolumn{1}{c|}{WV Enrichment} & \multicolumn{1}{c|}{Error Reduction} \\ \hline  
MNB                                      & 3 & 3 & 0.625                      & 0.641                      & 4.28\%                       & 0.371		& 0.398		& 4.23\%		                  \\ \hline
SVM                                      & 5 & 1 & 0.678                      & 0.681                      & 0.91\%                       & 0.539		& 0.554		& 3.32\%		\\ \hline 
\end{tabular}
%
\centering \small
\caption{\small DRC Dataset (2,159 instances, 64 Classes): Domain Specific Word Vector.}
\label{tab:drcfull}

\begin{tabular}{|l|c|c||r|r|r||r|r|r|}
\hline
\multicolumn{3}{|l||}{} & \multicolumn{3}{c||}{\textbf{Micro Recall}} & \multicolumn{3}{c|}{\textbf{Macro Recall}} \\ \hline 
 & \(n\) & \(k\) & \multicolumn{1}{c|}{Baseline} & \multicolumn{1}{c|}{WV Enrichment} & \multicolumn{1}{c||}{Error Reduction} & \multicolumn{1}{c|}{Baseline} & \multicolumn{1}{c|}{WV Enrichment} & \multicolumn{1}{c|}{Error Reduction} \\ \hline  
MNB                                      & 3 & 3 & 0.300                      & 0.337                      & 5.36\%                       & 0.115		& 0.144 		& 3.26\%		                  \\ \hline
SVM                                      & 3 & 3 & 0.375                      & 0.387                      & 1.93\%                       & 0.215		& 0.239		& 2.99\%		\\ \hline 
\end{tabular}
%
\centering \small
\caption{\small DRC Dataset (2,159 instances, 64 Classes): Top-3 Recommendation with Domain Specific Word Vector.}
\label{tab:drcfull3}

\begin{tabular}{|l|c|c||r|r|r||r|r|r|}
\hline
\multicolumn{3}{|l||}{} & \multicolumn{3}{c||}{\textbf{Micro Recall}} & \multicolumn{3}{c|}{\textbf{Macro Recall}} \\ \hline 
 & \(n\) & \(k\) & \multicolumn{1}{c|}{Baseline} & \multicolumn{1}{c|}{WV Enrichment} & \multicolumn{1}{c||}{Error Reduction} & \multicolumn{1}{c|}{Baseline} & \multicolumn{1}{c|}{WV Enrichment} & \multicolumn{1}{c|}{Error Reduction} \\ \hline  
MNB                                      & 3 & 3 & 0.494                      & 0.545                      & 10.05\%                       & 0.212		& 0.256 		& 5.59\%		                  \\ \hline
SVM                                      & 3 & 3 & 0.590                      & 0.611                      & 5.06\%                       & 0.357		& 0.396		& 6.07\%		\\ \hline 
\end{tabular}
\end{table*}

\subsubsection{OHSUMED Dataset}
Our second evaluation considered the evaluation of medical academic article references contained in the OHSUMED dataset \cite{HershBLH94}.
This dataset has been widely used in previous publications \cite{Joachims98,GabrilovichM06,WangD08}.  
We consider a subset of this dataset that contains references which only contain a title and no abstract. 
This particular problem setting has previously been considered by Gabrilovich and Markovitch \shortcite{GabrilovichM06} who used Wikipedia to enrich an SVM classifier.

The full OHSUMED dataset contains 348,566 references from MEDLINE covering 270 medical journals from the period 1987-1991.
Each article reference in this dataset is tagged with multiple human assigned MeSH indexing terms. 
We extract a subset of this dataset relating to the article references which have been assigned to either the first or second level of 21 MeSH Disease Categories. 
This produces a dataset of 44,479 references of which 12,752 references only contain a title with no abstract. 
We perform a 10x10-fold cross validation on this set of 12,752 references with the task of predicting which of the 21 Mesh Disease Categories each article reference belongs to. 
On average each article reference is labelled by 1.1 categories, however, to simplify this problem we configure out classifiers to only produce a single classification label and we consider an article reference to be correctly classified
if the linear model correct predicts one of its correct labels. 
This subset dataset contains 5,899 unique words and the mean length of each title is only 9 words. 
For our Word Vector enrichment process we train Word Vector models on the larger dataset of 44,479 article references using both the title and the abstract text (where available) under the same hyper-parameter configuration as the Reuters dataset. 

Table \ref{tab:ohsumedfull} shows the mean results of the 
cross validation where the full text of each title is used as input to the linear classifier. 
The baseline results for the SVM classifier again shows much higher performance than the baseline MNB result. 
When we enrich the bag-of-words model, we get a gain in both micro and macro recall for the MNB classifier. 
For the SVM classifier we get a much smaller gain in micro recall, however, the results are statistically significant \((p < 0.0005)\).

\subsubsection{ICG DRC Dataset}
This dataset contains “text snippets” extracted from 15 International Crisis Group (ICG) reports on the DRC
during the period 2002--2006.
To construct the dataset, a domain
expert read 8,836 sentences across the 15 reports, extracted 2,159 text snippets which were then each given one of 64 class labels.
This dataset contains 3,366 unique words and the mean length of each text snippet is 25 words.
Previous work has shown this dataset is very challenging for state of the art classification algorithms \cite{heap2017adma}.
We train a domain specific Word Vector model over the full text of all the sentences in the 15 ICG reports. 

Table \ref{tab:drcfull} shows that our Word Vector enrichment approach is able to improve the recall of both the MNB and SVM classifiers. 
All of these results are statistically significant \((p < 0.00005)\).
However, despite the improvement with the enrichment process, the linear classifiers still perform very poorly on this dataset. 
This is reflective of the difficultly in using this dataset for fully automated machine coding \cite{heap2017adma}. 
An alternative treatment of this dataset is to configure the classification algorithms to produce a ranked top-3 predictions of the class label 
which a human ``coder'' can then select from (cf. \cite{LarkeyC96}). 
The results of configuring the classifiers in this manner are presented in Table \ref{tab:drcfull3} and show that the enrichment method performs very 
strongly in this type of classification problem. 

Overall, our evaluation across three domain specific datasets has demonstrated that our bag-of-words enrichment with Word Vectors process improves the recall of linear classifiers in multi-class 
short text classification problems. 
A key finding in this work has been that using a small domain specific Word Vector model for enrichment is more effective at 
improving recall than a much larger general model which has been used in many previous works. 
A further benefit of this process is minimal computation required
to enrich texts, this is a consequence of the process only enriching unlabelled texts and therefore requires no retraining of existing classification models.

\section{Related Work}
A number of approaches for transforming the bag-of-words model or modifying linear models to increase the classification accuracy have been proposed in previous works. 
These approaches generally involve: 
1) Reducing the dimensionality of models by removing stop words \cite{MansuyH06}, stemming terms \cite{Porter80} and removing the least frequent words \cite{YangP97};
2) Transforming word frequencies by TF-IDF and other measures to identify the key terms in particular classes \cite{joachims1997probabilistic}; 
3) Limiting classes to a maximum vocabulary size through methods such as mutual information metrics \cite{DumaisPHS98}; 
4) Using dictionaries \cite{MavroeidisTVTW05,MansuyH06,Scott98textclassification} and encyclopaedias \cite{StrubeP06,WangD08} to find synonyms for rare terms. 
However, the names of people and organisations are unlikely to occur in dictionaries \cite{LuoCX11,MansuyH06} and may be used rarely in training text. 

These text transformation techniques can have a large influence over the accuracy of the classification models \cite{DumaisPHS98,YangP97}. 
However, the effectiveness of each method can vastly vary depending on the classification domain and problem, the dataset and the linear classifier. 
As such, there is no generally agreed best practice approach for addressing these issues. 
With regards to short text classification, any approach which reduces the number of words in the dataset or limits classes to their most frequent/relevant words are likely to reduce the classification accuracy as many words will lose any ability
to contribute to the classification of a text. 


\section{Conclusion}
In this paper we developed a method of enriching
bag-of-words representations for short text classification. 
We use a Word Vector model trained on unlabelled domain-related text to
capture semantic and syntactic relations between words. 
With the Word Vector model we locate words in the vector space which
may occur rarely in a set of labelled training examples, then use the
neighbour words to enrich the bag-of-words vector to be classified. 
This enriched representation allows linear classifiers to use the
class information of more words to enhance classification prediction. 
A key benefit of the approach is that it does not
require any change to the training of linear classification models. 
Combining the information in the unsupervised Word Vector
model with a supervised linear model improves micro and
macro recall over baseline on difficult multi-class problems,
although the potential of CNNs for these should be investigated further.

\section*{Acknowledgment} 
This work was supported by Data to Decisions Cooperative Research Centre.
We are grateful to Josie Gardner for labelling the ICG DRC dataset.
 
\bibliographystyle{aaai}
\bibliography{refs}

\end{document}